\documentclass[letterpaper]{article} % DO NOT CHANGE THIS
\usepackage[submission]{aaai23}  % DO NOT CHANGE THIS
\usepackage{times}  % DO NOT CHANGE THIS
\usepackage{helvet}  % DO NOT CHANGE THIS
\usepackage{courier}  % DO NOT CHANGE THIS
\usepackage[hyphens]{url}  % DO NOT CHANGE THIS
\usepackage{graphicx} % DO NOT CHANGE THIS
\usepackage{natbib}  % DO NOT CHANGE THIS AND DO NOT ADD ANY OPTIONS TO IT
\usepackage{caption} % DO NOT CHANGE THIS AND DO NOT ADD ANY OPTIONS TO IT
\usepackage{algorithm}
\usepackage{algorithmic}

\usepackage{newfloat}
\usepackage{listings}
\DeclareCaptionStyle{ruled}{labelfont=normalfont,labelsep=colon,strut=off} % DO NOT CHANGE THIS
\floatstyle{ruled}
\newfloat{listing}{tb}{lst}{}
\floatname{listing}{Listing}
\title{TIER-A: Denoising Learning Framework for Information Extraction}

\author {
    Yongkang Li,\textsuperscript{\rm 1}
    Ming Zhang, \textsuperscript{\rm 1}
}
\affiliations {
    \textsuperscript{\rm 1} Peking University\\
    liyongkang@pku.edu.cn, mzhang\_cs@pku.edu.cn
}

\usepackage{bibentry}
\begin{document}

\maketitle

\begin{abstract}
With the development of deep neural language models, great progress has been made in information extraction recently. However, deep learning models often overfit on noisy data points, leading to poor performance. In this work, we examine the role of information entropy in the overfitting process and draw a key insight that overfitting is a process of overconfidence and entropy decreasing. Motivated by such properties, we propose a simple yet effective co-regularization joint-training framework \textbf{TIER-A}, \textbf{A}ggregation Joint-training Framework with \textbf{T}emperature Calibration and \textbf{I}nformation \textbf{E}ntropy \textbf{R}egularization. Our framework consists of several neural models with identical structures. These models are jointly trained and we avoid overfitting by introducing temperature and information entropy regularization.
Extensive experiments on two widely-used but noisy datasets, TACRED and CoNLL03, demonstrate the correctness of our assumption and the effectiveness of our framework.
\end{abstract}

\section{Introduction}

Significant advances in information extraction (IE) tasks have been achieved in recent years with the prosperity of deep neural models\cite{GRU}\cite{transformer}\cite{BERT}\cite{luke}. Despite the gratifying progress, deep neural models overfit easily on noisy datasets, leading to poor performance\cite{tacred-revisited}. Unfortunately, in large-scale text datasets, whether annotated manually or automatically, there will inevitably be a large number of incorrectly labeled data points. This problem exists in most widely used benchmarks, such as TACRED\cite{tacred} and CoNLL03\cite{conll03}, influencing the performance of SOTA methods\cite{tacred-revisited}. Hence, it is of vital importance to develop a noise-robust training method for related tasks.

At present, few works have been done on the noise-robust model in IE tasks. Existing work mainly focused on weakly supervision or distantly supervision\cite{ds-begin}\cite{mil1}\cite{mil2}\cite{mil3}\cite{mil4}. Such methods rely on multi-instance learning and can hardly be adapted to supervised settings. Another line is using reinforcement learning to find mislabeled data in datasets\cite{dsrl1}\cite{dsrl2}. However, such methods are highly related to the noisy rate of the validation set, as they are using the change of F1 score on the validation set as the rewards for agents. 
In supervised setting, a representative work is CrossWeight \cite{crossweight}, which trains multiple independent models on different folds of training data. Data points with inconsistent model predictions will be marked as noise points and their training weights will be reduced. Although this method is effective, it needs to train tens of model copies. Not only is it inefficient, but also difficult to be used in large language models commonly used in natural language processing. Another work is NLLIE \cite{NLLIE}. NLLIE is a joint-training framework, which uses the average of several independent models' predictions as a soft label to prevent model overfitting. Although this method achieves excellent performance, their aggregation upon predictions failed to control the prediction confidence, resulting in overfitting during the later stage of training. According to our research, denoising learning in supervised IE tasks has not been well studied.

In this paper, we propose a compatible noise-robust joint-training framework, as illustrated in Fig \ref{fig2}. Our model is motivated by former studies\cite{guo2017calibration}\cite{DCN}, which show that deep neural networks tend to make absolute confident predictions. The maximum value of softmax output will be closer and closer to 1 during the process of overfitting. The overfitting and overconfidence in the seen data will hurt the generalization ability of the models. Hence, we introduce temperature calibrated aggregation and information entropy regularization to mitigate the overconfidence in predictions. Our framework consists of two or more independent neural models with different parameter initialization. All models are encouraged to give predictions, which are similar to each other and have greater information entropy. By introducing temperature calibrated task loss, temperature calibrated co-regularization loss, and maximum information entropy regularization, our framework effectively prevents models from overfitting. 

In this paper, experiments are carried out on the deep language models including Albert\cite{albert}, Bert\cite{BERT} and LUKE\cite{luke}, and the datasets of TACRED\cite{tacred} and CoNLL03. The experimental results show that the proposed framework has excellent noise robust training ability and can stably and significantly improve the model performance. The contributions of this paper are summarized as follows:
\begin{itemize}
\item [1.] This paper proposes a noise-robust training framework for information extraction tasks. This framework not only has satisfying performance but also good interpretability.
\item [2.] The method proposed in this paper is independent of the information extraction model, which means that it can be directly applied to any advanced information extraction model.
\item [3.] Extensive experiments on named entity recognition and relationship extraction benchmarks show that the proposed training framework can achieve significant improvement with many SOTA models, and outperforms existing methods.
\end{itemize}

\section{Related Work}

In this section, we will mainly discuss two lines of related work. As each of them has a large number of works, our summary will be brief and selected.

\subsection{Distant Supervision} Distant Supervision\cite{ds-begin} combines rich unlabeled data with existing knowledge base via heuristic method. In this way, we will get a large amount of training data containing noise. In order to reduce the impact of data noise on model training, a main method is multi-instance learning \cite {mil1} \cite{mil2} \cite {mil3}\cite {mil4}. Their motivation is that given a batch of noisy data, at least one data is correctly labeled. PCNN \cite{mil2} uses the instance most consistent with the label of instance bags as training data, where the label of the bag is positive if and only if at least one instance in the bag is predicted to be positive. However, in Supervised learning, most of the data are correctly labeled, it will give us no benefits if we give up tens of data to obtain a promising clean data point. Another direction is to use reinforcement learning agents to relabel the noise data in the training set\cite{dsrl1}\cite{dsrl2}. By relabeling the training set and using the F1 score in the validation set as policy-based rewards, agents do learn how to find mislabeled data points and how to relabel them. However, this method is highly dependent on a perfectly labeled validation set, which is expensive to get in information extraction tasks.

\subsection{Supervision}

In the field of computer vision, there are many related research directions, such as mutual learning\cite{dml}\cite{multitaskdml}\cite{mutualnet}, noise filtering layer\cite{nf1}\cite{nf2}, robust loss function or regularization\cite{robustloss}\cite{robustloss2} and sample selection\cite{ss0}\cite{ss1}\cite{ss2}. The above works are very inspiring to our research. Deep mutual leaning\cite{dml}\cite{multitaskdml}\cite{mutualnet} is a joint-training framework, which consist of several deep models. By using Kullback-Leibler Divergence, models' predictions are encouraged to approach the average of others to prevent overfitting. Noise filtering layer\cite{nf1}\cite{nf2} uses a probability transfer matrix to simulate the process from real labels to noise data. Robust loss function method reduces the influence of noise labels on the robustness of the model by designing an appropriate loss function such as Symmetric Cross Entropy\cite{robustloss2}. Sample selection\cite{ss0}\cite{ss1}\cite{ss2} assumes that data instances with large training losses are noise points and will remove them from training. Although this does reduce the impact of noise points, it will also eliminate some long tail classes from the dataset.

Under the supervision of natural language processing, there are only a few related researches. The representative works include CrossWeight \cite{crossweight} and NLLIE\cite{NLLIE}. Crossweight trains multiple independent models on folds of training data. Data points with inconsistent predictions will be regarded as noise points and be assigned with lower training weights. Although this method is effective, it needs to train tens of model copies, which is not only inefficient but also difficult to be used in large language models commonly used in natural language processing. NLLIE is a joint-training framework, which uses a similar strategy to deep mutual learning. Although this method achieves excellent performance, their aggregation upon predictions failed to control the prediction confidence, resulting in overfitting during the later stage of training. 

\section{Preliminary}

The relation extraction task aims to determine whether a given relationship exists for the specified entity pair in the text. Specifically, the goal of the relationship extraction task is to train a classifier $f $, predict the relationship between $(x, e_s, e_o) $and $(e_s, e_o) $for each given triplet. Where $x $is the given sentence, $e_s$ and $e_o$represent the potential subject and object respectively. Following the previous work \cite{re1} \cite{re2} \cite{mtb}, this paper formalizes this task as a sentence classification problem, and uses an entity mask to deal with sentences. For example, $\textit {jobs founded Apple}$ will be formulated to $\textit{[CLS] [subject-person] founded}$ $\textit{[object-company] [SEP]}$, where [CLS] and [SEP] marks the beginning and ending of a sentence. We will use [CLS] representation to get the prediction over the relations between entities.

The named entity recognition task aims to find the location of named entities in text and predict their classes. Following the work of Devlin et al. \cite{BERT}, this paper also formalizes this task as a token classification problem. Specifically, given a text sentence $\textbf {X} = \{x_{i}\}_{i=1}^N $, it will be instantiated into a sub-token sequence $\textbf {Z} = \{z_{ij}\}_{i=1}^{N} $, where $Z_ {ij}$ is the $j$-th sub-token of $x_i$. The representation of the first sub-token of each word will be used as the input of classifiers to determine whether the word is a named entity or not and which type of named entity it belongs to.

\section{Method}

\begin{figure*}[h]
\centering
\includegraphics[width=2\columnwidth]{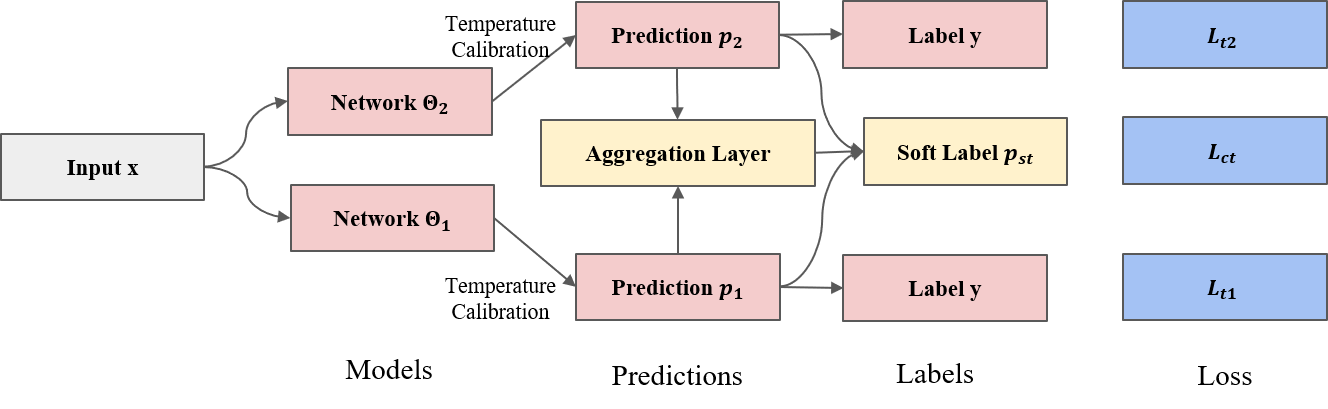} % Reduce the figure size so that it is slightly narrower than the column. Don't use precise values for figure width.This setup will avoid overfull boxes.
\caption{Overview of TIER}.
\label{fig2}
\end{figure*}

The overview of TIER-V model is shown in Fig \ref{fig2}.

\subsection{Learning Process}
 Our framework consists of $n$ ($n\ge 2$) copies of the task-specific model with different initialization. For models built upon pre-trained language models, only the external parameters (e.g., those of a downstream classifier) are randomly initialized, while the pre-trained parameters are the same. To balance the efficiency and effectiveness, we use two model copies ($n=2$) in this paper. Once initialized, we will have a two-phase training process. The training algorithm is described in Alg. \ref{alg:algorithm}.

The first phase is the warm-up stage. In this stage, the joint-training loss function is deactivated. Only is the temperature calibrated independent loss function of each model used to complete the training of $n$ models. As the model converges much faster on the well-labeled data than on the noisy data\cite{arpit2017closer}, a sufficient independent training phase can help the model quickly fit the main data pattern.

The second phase is the joint-training stage. In this stage, the model will conduct training according to both independent training loss and well-designed joint-training loss function. Specifically, the temperature calibrated predictions of $n$ models are aggregated through the aggregation layer, whose result will be used as a soft label to encourage models to have a consistent prediction.

\begin{algorithm}[tb]
\caption{TIER-A:Training Algorithm}
\label{alg:algorithm}
\textbf{Input}: dataset $D$ \\
\textbf{Parameter}: hyperparameters $T,\alpha,\beta,\gamma,n,S$, base model parameters $\{\Theta_i\}_{i=1}^n$\\
\textbf{Output}: An emsemble classifier $F$
\begin{algorithmic}[1] %[1] enables line numbers
\STATE Let $t=0$.
\FOR{s=1,2,3,...,S}
\STATE Sample a batch $B$ from $D$
\STATE Calculate independet training loss $\{L_{t}^{i}\}_{i=1}^n$ by Eq. 5.
\STATE $L_{ind}= \frac 1 n\sum_{i=1}^nL_{t}^{i}$
\IF{$s\le S*\gamma$}
\STATE Update $\{\Theta_i\}_{i=1}^n$ w.r.t. $L_{ind}$
\ELSE 
\STATE Calculate the soft label $p_st$ by Eq. 3.
\STATE Calculate the joint-training loss $L_{ct}$ by Eq. 4.
\STATE $L=L_{ind}+\beta L_{ct}$
\STATE Update $\{\Theta_i\}_{i=1}^n$ w.r.t. $L$
\ENDIF
\ENDFOR

\STATE \textbf{return} $\{\Theta_i\}_{i=1}^n$
\end{algorithmic}
\end{algorithm}

\subsection{Optimization Objective}
Our optimization objective function consist of two parts. The second part is independent training loss. 

The first part is joint-training loss. All logits output will be gathered to go through the aggregation layer, which will generate a soft label according to the specified evaluation method.

\begin{equation}
    p_{st} = Agg(l_1,...,l_n;T)
\end{equation}

where $n$ is the number of models used in the framework and $Agg(·)$ is the aggregation function determined by the aggregation method. $T$ is the temperature hyper-parameter. Temperature calibration was originally proposed by Hinton et al\cite{KD} to distill knowledge from neural models. We apply temperature calibration to transform the prediction as

\begin{equation}
    p_{i} = Softmax(l_i/T)
\end{equation}

Temperature $T$ decrease the entropy when $0<T<1$. As $T \rightarrow 0$, the distribution collapses to a point mass, leading to a higher certainty. By setting low temperatures, models can greatly reduce the impact of noisy labels. For example, assume that we have a label (1,0). By setting t=0.2, a logit output of (0.6,0.4) will lead to a prediction of (0.73,0.27) instead of (0.55,0.45). The latter will draw a great loss when dealing with wrong labels, leading to a quick rush in the overfitting phase. For data following the main pattern(clean data), a low temperature will curb the trend of overconfidence. When dealing with noisy data points, a lower temperature will encourage the logit output to form a stalemate between true class and wrong label. By curbing the trend of overconfidence in both noisy data and clean data, deep models will have more time to learn long-tail classes before being convinced by noisy data points, and thus have a better performance. 

Former works\cite{NLLIE}\cite{arpit2017closer} indicate that deep models can easily fit well-labeled data but need more steps to fit the noisy data. Thus on clean data, models can easily reach an agreement, while on noisy data their predictions are usually inconsistent. By introducing average aggregation, noisy data will arouse great agreement loss, which can reduce their influence on optimization. 

\begin{equation}
    Agg(l_1,...,l_n;T)= \frac 1 n \sum_{i=1}^n p_i
\end{equation}

The joint-training loss is as follows
\begin{equation}
    L_{ct} = \frac 1 n\sum_{i=1}^n KL(p_{st},p_i)
\end{equation}
where KL is the KL divergence from $p_i$ to $p_{st}$.

The second part is independent-training loss.
\begin{equation}
    L_{t}^{i} = CE(p_i,y)+\alpha NIE(p_i)
\end{equation}

where CE is the CrossEntropy loss, NIE is the negative information entropy of distribution $p_i$, and $p_i$ is the temperature-calibrated prediction of model $i$. By introducing negative information entropy regularization(NIE), we encourage the model to maintain reasonable uncertainty when giving a prediction and thus prevent the tendency of overfitting.

The objective function will be

\begin{equation}
    L = \frac 1 n\sum_{i=1}^n {L_{t}^{i}}+\beta L_{ct}
\end{equation}
Expand this expression, we will have
\begin{equation}
    L =\frac 1 n\sum_{i=1}^n \{ CE(p_i,y)+ \alpha NIE(p_i)+\beta KL(p_{st},p_i)\}
\end{equation}

\subsection{Prediction}
Motivated by ensemble learning methods, we will use weighted voting when making predictions. Intuitively, the logit outputs generally reveal the confidence of each model for each class, we design our confidence soft vote classifier as
\begin{equation}
    r_p = \arg \max l^{r}_c
\end{equation}
where $r_p$ is our predicted class, $r \in [0,1,...,K]$, $K$ is the number of classes. $l_c$ is the soft vote prediction as
\begin{equation}
   l_c = \sum_{i=1}^n l_i/||l_i||
\end{equation}
where $l_i$ is the logit output of model $i$, and n is the number of models used in our framework.

\begin{figure*}[h]
\centering
\includegraphics[width=2\columnwidth]{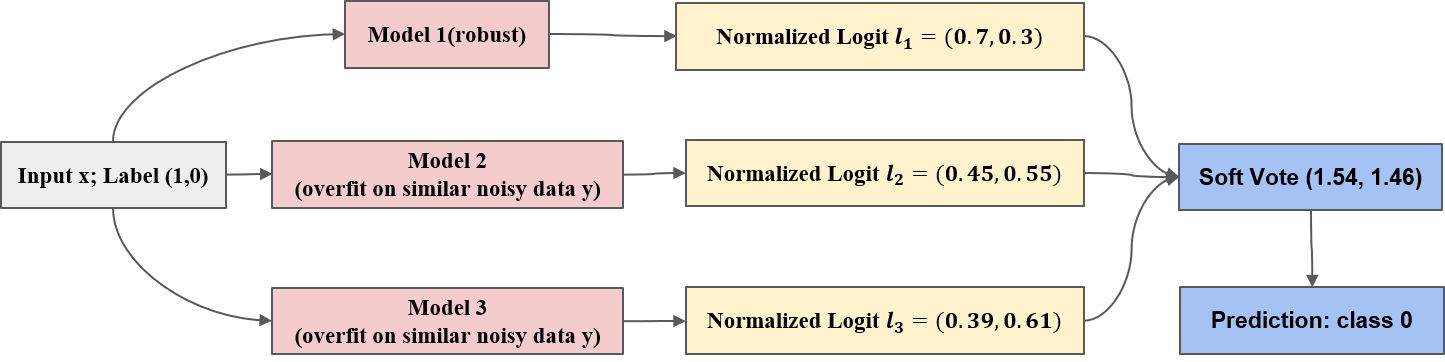} % Reduce the figure size so that it is slightly narrower than the column. Don't use precise values for figure width.This setup will avoid overfull boxes.
\caption{A Toy Example}.
\label{example}
\end{figure*}

According to our design of training objective function, models are supposed to have sufficient uncertainty when dealing with data points that conflict with the main data pattern. Given a noisy input, overfitting models will give a logit output where the noisy label only exceeds slightly over the true class. Thus it only takes one robust model which can correctly classify the noisy data points to enable the joint-classifier to make a proper prediction. A toy example is shown in Figure \ref{example}. The framework consists of three model copies to handle a binary classification problem. For a given input $x$ belonging to class 0, two of the base models have overfitted on similar noisy data group y while one remains robust. Although convinced by noisy data, due to the conflict of main data pattern, temperature calibration, and maximum information entropy regularization, the logits from overfitting models are in high uncertainty so that the prediction of a robust model can easily dominate in the voting stage, leading to correct prediction.

Note that our key intuition for the soft vote is highly related to the magnification of temperature calibration, which assures that a model will give a prediction with high uncertainty on data points similar to noisy data even if it has been convinced by specific noisy data patterns. In this case, if there exists a robust model, its confidence in the main data pattern can easily occupy the dominant position in the final prediction.

This kind of soft vote may not always be useful for other joint-training frameworks such as our baseline NLLIE\cite{NLLIE}. In NLLIE's setting, their softmax function will not magnify the gap between different classes in logit output. Thus, if a model is confused by a noisy pattern, it will give predictions with pretty strong confidence. In this case, even if there exists a more robust model, doing the average on logit or prediction distribution still gives no guarantee for a proper prediction.

By introducing soft vote, our model can gain greater performance gain with more copies, which we will discuss in the experiments section.

\section{Experiment}
\subsection{Datasets}
Our experiments will be conducted on the TACRED and CoNLL03 datasets.
\begin{itemize}
\item \textbf{TACRED} dataset is a large-scale relational extraction dataset developed by the Stanford NLP group. It contains 106264 annotated text data, respectively belonging to 41 TAC KBP relation classes or annotated as $no\_ relation$.  Former work \citep{tacred-revisited} pointed out that there is a large group of noisy data points on it, and accordingly relabeled its validation set and test set.
\item \textbf{CoNLL03} is a manually annotated named entity identification data set. Recent work \cite{crossweight} pointed out that 5.38\% of the sentences in the data set had at least one word incorrectly labeled. And they provide a relabeled version of the test set.
\end{itemize}

The statistics of the data are shown in table \ref{dataset}. For this paper, we will mainly focus on the performance on relabeled validation and test set. For TACRED, as it has a complete relabeled evaluation set, we report the F1 score on both the original and relabeled validation and test set. For CoNLL03, which has a much lower noise rate, as its relabel version only has a relabeled test set, we only report the F1 score on the relabeled test set for simplicity and comparability.

\begin{table}[h]
	\centering  % 显示位置为中间

	%字母的个数对应列数，|代表分割线
	% l代表左对齐，c代表居中，r代表右对齐
	\begin{tabular}{|c|c|c|c|c|c|}  
		\hline  % 表格的横线
		Dataset & \#train & \#dev & \#test & \#classes & \% noise\\
		\hline
        TACRED &68124 &22631 &15509 &42 &6.62\\
        \hline
        CoNLL03 &14041 &3250 &3453 &9 &5.38\\
		\hline
	\end{tabular}
	\caption{Data Statistics}  % 表格标题
	\label{dataset}  % 用于索引表格的标签
\end{table}

\subsection{Model Configurations}

We implement all base models (ALBERT-base, BERT-base, BERT-large, and LUKE) based on Huggingface’s Transformers \cite{wolf2019huggingface}. All models are optimized with Adam\cite{adam}. For the most part of the experiments, we use a batchsize of 64 and a learning rate of 6e-5 for TACRED and 1e-5 for CoNLL03. Considering the size of model and capacity of our device, we may scale down batchsize and learning rate for certain base models. We finetune the TACRED model for 5 epochs and the CoNLL03
model for 50 epochs. And we choose the best checkpoint based on the F1 score on the validation set. For TACRED, we use F1 score on the relabeled validation set. For CoNLL03, as there is no relabel version of validation set, we use F1 score on original validation set instead to choose a model checkpoint. We tune $ \alpha$ from $\{1e-5, 1e-4,1e-3, 1e-2, 5e-2\}$, $\beta$ from $\{5,10\}$and tune $T$ from $\{0.2, 0.3, 0.5, 0.8\}$. The median F1 of 5 runs using random seeds will be reported.

For efficiency, we use two model copies (n = 2) in the following experiments. We will discuss the situation of more copies in later section, where our improvement can be even more significant.

\subsection{Base Model}
This paper tests the effectiveness of the proposed method on the following SOTA models.
\begin{itemize}
\item \textbf{ALBERT} is a lightweight improved version of Bert. The ALBERT-base-v2 version was used in the experiment in this paper.
\item \textbf{BERT} is a transformer-based pre-training language model. Both base and large versions were considered in the experiment in this paper.
\item \textbf{LUKE} is another transformer-based language model, which has been pretrained on large-scale text corpus and knowledge graph. It achieves SOTA performance on various tasks, including RE and NER.

\end{itemize}

\subsection{Main Results}
The results in TACRED and CoNLL03 is shown in Tab \ref{mTacred} and Tab \ref{mCoNLL03}. The experiments are divided into several groups. Each group focuses on one pre-trained language model, including four results: the results from a model without any joint-training method(marked as X-Ori), that with our baseline NLLIE(marked as X-CR) ,and that with our method(marked as X-TIER, X-TIERV). As our baseline NLLIE uses only one model's prediction for evaluation, to make a comparable result, we will report two versions of our performance, X-TIER for not using vote and X-TIERV for using logits vote.

In TACRED, we consistently outperform baseline even without soft vote, which indicates that our training framework can lead to more robust models. When introducing soft vote, we consistently enhance the existing model for 3.1\% to 3.8\% and outperform the baseline by from 0.7\% to 1.6\%. In lite models such as Albert(12M) and BERT-base(108M), our improvement is very significant, but for BERT-large(334M) and LUKE(253M) our improvement is relatively wispy. 

In CoNLL03, where data is much cleaner than that in TARCED, our model can still provide a consistent performance gain. 

\begin{table*}[h]
	\centering  % 显示位置为中间

	%字母的个数对应列数，|代表分割线
	% l代表左对齐，c代表居中，r代表右对齐
	\begin{tabular}{|c|c|c|c|c|}  
	    \hline
        Model &Dev F1 &Test F1 &Dev-rev F1 &Test-rev F1\\
        \hline
        AlBERTbase-Ori &68.527&68.524&76.337&77.390\\
        AlBERTbase-CR &70.226&69.792&79.149&79.607\\
        AlBERTbase-TIER &70.817&70.742&80.063&80.704\\
        AlBERTbase-TIERV &70.523&70.608&80.468&81.001\\
        \hline
        BERTbase-Ori &69.424&69.696&77.003&77.392\\
        BERTbase-CR &71.296&70.579&79.710&79.704\\
        BERTbase-TIER &72.078&71.867&80.521&81.073\\
        BERTbase-TIERV &71.923&72.29&80.944&81.277\\
        \hline
 
        BERTlarge-Ori &71.43&70.603&78.684&78.324\\
        BERTlarge-CR &72.938&73.085&81.265&81.814\\
        BERTlarge-TIER &72.974&72.854&81.628&81.882\\
        BERTlarge-TIERV &73.392&72.973&81.717&82.402\\
        \hline
        LUKE-Ori &70.284&70.102&79.300&79.551\\
        LUKE-CR  &71.667&71.857&82.108&82.378\\
        LUKE-TIER &71.315&71.481&81.712&82.735\\
        LUKE-TIERV &71.496&71.586&82.312&83.119\\
        \hline
	\end{tabular}
	\caption{Main Result on TACRED}  % 表格标题
	\label{mTacred}  % 用于索引表格的标签
\end{table*}

\begin{table}[h]
	\centering  % 显示位置为中间

	%字母的个数对应列数，|代表分割线
	% l代表左对齐，c代表居中，r代表右对齐
	\begin{tabular}{|c|c|}  
	    \hline
        Model &Test-rev F1\\
        \hline
        AlBERTbase-Ori &90.63\\
        AlBERTbase-CR &91.14\\
        AlBERTbase-TIER &91.20\\
        AlBERTbase-TIERV &91.63\\
        \hline
        BERTbase-Ori &92.95\\
        BERTbase-CR &93.20\\
        BERTbase-TIER &93.22\\
        BERTbase-TIERV &93.27\\
        \hline
        BERTlarge-Ori &93.25\\
        BERTlarge-CR &93.54\\
        BERTlarge-TIER &93.51\\
        BERTlarge-TIERV &93.57\\
        \hline
	\end{tabular}
	\caption{Main Result on CoNLL03}  % 表格标题
	\label{mCoNLL03}  % 用于索引表格的标签
\end{table}

\subsection{Ablation Study}
To verify the effectiveness of each component in our framework, we have tested several combinations of them and report the median F1 score of five runs on relabeled test set. The results are shown in Tab \ref{tier}.
\begin{table}[h]
	\centering  % 显示位置为中间

	%字母的个数对应列数，|代表分割线
	% l代表左对齐，c代表居中，r代表右对齐
	\begin{tabular}{|c|c|} 
	\hline
	    Combination&test-rev F1\\
	    \hline		
        Albert+A			&	79.607\\
        Albert+A+T		&	80.513\\
        Albert+A+TIER	&	80.704\\
        \hline
        BERT+A			&	79.704\\
        BERT+A+T		    &	81.044\\
        BERT+A+TIER	    &	81.073\\
	    \hline
        LUKE+A	        &	82.378\\
        LUKE+A+T			&	82.629\\
        LUKE+A+TIER		&	82.735\\
        \hline
	\end{tabular}
	\caption{ F1 score (\%) of different combinations of components on the relabeled test set of TACRED}  % 表格标题
	\label{tier}  % 用于索引表格的标签
\end{table}

The experiments are divided into three groups, each has four lines. X+A represents a joint-training framework with average aggregation, X+A+T introduces temperature calibration and X+A+TIER introduces both temperature calibration and information entropy regularization. Note that we have already shown X+A+TIER in Tab \ref{mTacred} as X-TIER.

The temperature calibration and confidence soft vote contribute the most part of the performance gain and IER's contribution is relatively small.

\subsection{Noise-robust test}To illustrate the noise-robustness of our method, it is vital for us to report our performance on dataset with different noise rates. Thus we further evaluate our framework on several noisy versions of TACRED. We randomly flipping 10\%, 30\%, 50\%,
70\%, or 90\% labels in the training set of TACRED. Then we use those noisy training sets to train RE models(we use BERT-base for this experiment) and evaluate them on the relabeled test set of TACRED. 

We report the median F1 score of five runs on relabeled test set. Results are shown in Tab \ref{Noisy Rate}. Note that the original BERT only uses clean data subset, deleting all flipped data points. 
\begin{table*}[h]
	\centering  % 显示位置为中间

	%字母的个数对应列数，|代表分割线
	% l代表左对齐，c代表居中，r代表右对齐
	\begin{tabular}{|c|c|c|c|c|c|}  
	    \hline
	    Model &10\% &30\% &50\%  &70\% &90\% \\
	    \hline
	    BERT-CR&79.149&77.193&73.216&62.283&21.353\\
        BERT-Tier&80.158&77.946&74.469&63.652&27.164\\
        \hline
        Bert(No noise)&76.287&75.276&73.429&68.666&55.213\\
        \hline
	\end{tabular}
	\caption{ F1 score (\%) under different noisy rates on the relabeled test set of TACRED}  % 表格标题
		\label{Noisy Rate}  % 用于索引表格的标签
\end{table*}
It is obvious that our method significantly outperforms both our baseline and BERT trained with the clean data subset when the noisy rate is no greater than 50\% percent, demonstrating the noise-robustness of our method.

\subsection{Extra Copies}
The main results show the promising improvement via our method with two model copies. Intuitively, using more copies may achieve higher performance gain. As shown in Tab \ref{N-model}, we further explore the performance on TACRED when using three and more BERT-base copies. We report the median F1 score of five runs on relabeled test set. 

The performance of our baseline NLLIE\cite{NLLIE} is marked as BERT-CR, the performance for a single model trained in our framework is marked as BERT-TIER while the performance of the soft vote from all joint-training models are marked as BERT-TIERV.
\begin{table}[h]
	\centering  % 显示位置为中间

	\begin{tabular}{|c|c|c|c|} 
	\hline
	    n-model&2&3&4\\
	    \hline
	BERT-CR  &79.7 &79.5 &79.8\\
    BERT-TIER    &81.1&81.4&81.3\\
    BERT-TIERV    &81.3&81.7&81.4\\
        \hline
	\end{tabular}
	\caption{F1 score (\%) under different model copies on the relabeled test set of TACRED}  % 表格标题
	\label{N-model}  % 用于索引表格的标签
\end{table}
On BERT-base, increasing the number of copies slightly improves each model's performance(from 81\% to 81.35\%). This indicates the gain from prediction aggregation will reach a soft upper limit when use three copies. However, by using soft vote, extra performance gain from 0.5\% to 1\% can be achieved when adding more copies. 

We also observe a decline of performance when using four copies(n=4). This is mainly due to that models are harder to reach an agreement if there are too many of them. This problem can be solved by simply increasing the number of training epochs or training steps.

Note that as models in our framework can be trained in parallel, increasing the number of models does not necessarily increase training time, but cost more computational resources.

\section{Conclusions}
In this paper, we propose a joint-training framework to reduce the influence of noise data when training deep language models in information extraction tasks. The proposed method is composed of several models with different initialization. Specifically, we designed a joint-training loss and an independent training loss with temperature calibration, which encourage models to give similar predictions with sufficient uncertainty. Extensive experiments on NER and RE benchmarks show that our model achieves significant improvement on various IE models.

% \subsection{evaluation method}
% Tab \ref{evaluation}
% \begin{table}[h]
% 	\centering  % 显示位置为中间
% 	\caption{evaluation method}  % 表格标题
% 	\label{evaluation}  % 用于索引表格的标签
% 	%字母的个数对应列数，|代表分割线
% 	% l代表左对齐，c代表居中，r代表右对齐
% 	\begin{tabular}{|c|c|} 
% 	\hline
% 	    Combination&test-rev F1\\
% 	    \hline		
%         bert+minie+t    &81.173\\
%         \hline
%         bert+avg+tier=-0.05&81.073\\
%         \hline
% 	\end{tabular}
% \end{table}
\bibliography{anonymous-submission-latex-2023.bbl}
\end{document}